\documentclass[letterpaper, 10 pt, conference]{ieeeconf}  % Comment this line out if you need a4paper

\IEEEoverridecommandlockouts                              % This command is only needed if
\title{\LARGE \bf
Balance Between Efficient and Effective Learning: {Dense2Sparse} Reward Shaping for Robot Manipulation with Environment Uncertainty
}

\author{Yongle~Luo$^{*}$,
Kun~Dong$^{*}$,
Lili~Zhao$^{}$,
Zhiyong~Sun$^{}$,
Chao~Zhou$^{}$,
and Bo~Song$^{\#}$
\thanks{This work is supported in part by the National Natural Science Foundation of China (Grant No. 61973294) and Key Research and Development Plan of Anhui Province (Grant No.01904a05020086).}
\thanks{$^{*}$These authors are contributed equally.}
\thanks{Y.~Luo, K.~Dong, L.~Zhao, Z.~Sun and B.~Song are with Institute of Intelligent Machines, Hefei Institute of Physical Science, CAS, Hefei, 230031, China}
\thanks{C.~Zhou is with Institute of Plasma Physics, Hefei Institute of Physical Science, CAS, Hefei, 230031, China}
\thanks{$^{\#}$Corresponding author, email: {\tt\small songbo@iim.ac.cn}}
}

\usepackage{graphicx}
\usepackage{subfigure}
\usepackage{caption}
\usepackage{amsmath}%数学包
\usepackage{amssymb}%公式包
\usepackage{bm}
\usepackage{tabularx}
\usepackage{verbatim}
\usepackage{cite}
\usepackage{textcomp}
\usepackage{xcolor}
\usepackage{algorithmic}
\usepackage{comment}

\begin{document}
\bibliographystyle{IEEEtran}

\maketitle
\thispagestyle{empty}
\pagestyle{empty}

%%%%%%%%%%%%%%%%%%%%%%%%%%%%%%%%%%%%%%%%%%%%%%%%%%%%%%%%%%%%%%%%%%%%%%%%%%%%%%%%
\begin{abstract}

Efficient and effective learning is one of the ultimate goals of the deep reinforcement learning (DRL), although the compromise has been made in most of the time, especially for the application of robot manipulations. Learning is always expensive for robot manipulation tasks and the learning effectiveness could be affected by the system uncertainty. In order to solve above challenges, in this study, we proposed a simple but powerful reward shaping method, namely \emph{Dense2Sparse}. It combines the advantage of fast convergence of dense reward and the noise isolation of the sparse reward, to achieve a balance between learning efficiency and effectiveness, which makes it suitable for robot manipulation tasks. We evaluated our {Dense2Sparse} method with a series of ablation experiments using the state representation model with system uncertainty. The experiment results show that the {Dense2Sparse} method obtained higher expected reward compared with the ones using standalone dense reward or sparse reward, and it also has a superior tolerance of system uncertainty.
\end{abstract}

%%%%%%%%%%%%%%%%%%%%%%%%%%%%%%%%%%%%%%%%%%%%%%%%%%%%%%%%%%%%%%%%%%%%%%%%%%%%%%%%
\section{Introduction}

Deep reinforcement learning (DRL) has shown its amazing ability in real robotic manipulations ranging from object grasping~\cite{r1,r2}, ball playing~\cite{r3,r4}, door opening operations~\cite{r7} and assembly tasks~\cite{r5,r6}. Compared to conventional control methods, DRL utilizes a powerful nonlinear approximator to learn a mapping from state to action, which is more robust to the noise in the states for conducting complex tasks \cite{r5,r8,r9}. Most of the previous works in the DRL guided robot manipulations assumed that the agents are able to obtain accurate rewards during the training process~\cite{r1,r5}. However, in real practice, noise from the environment or the sensor itself could increase the system uncertainty. It also affects the learning effectiveness of the DRL~\cite{r10,r11,r12}, due to the error accumulation.\par

There are two major factors that affect the learning efficiency and effectiveness of DRL: reward shaping and policy. Recently, various learning polices have been intensively studied with great improvement on the performance~\cite{r13,r14,r15,r16,r17,r18}. However, only a few attempts have been made for improving reward shaping methods. In the DRL, the reward shaping mainly includes two schemes, the sparse reward and the dense reward~\cite{r19,r20,r21,r22,r23}. Sparse reward is a straightforward reward shaping method that makes the agent obtain reward only if the task or sub-task has been completed. In such a case, the agent is required to explore the environment without any reward feedback, until it achieves the target state. That makes the policies using sparse reward difficult to be optimized since there is little gradient information to drive the agent to the target~\cite{r24}. In addition, it usually takes a long period of time to get a successful trajectory with random actions in the high dimensional state. In order to solve the efficiency issue of the sparse reward, the dense reward was developed with continuous feedback to the agent to accelerate the learning~\cite{r19,r21,r22,r23}. Although the dense reward-based DRL usually has a faster learning speed compared to the sparse one, the learning effectiveness highly depends on the accuracy of reward, which can be affected by the noise and disturbance from the environment.\par

It is noted that at the beginning of the learning process using the sparse reward, the learning converges relatively slow due to the lack of global information and successful experience. However, when enough successful data have been stored into the buffer, the sparse reward can lead to better training results~\cite{r2,r25}. In the opposite, the dense reward can learn a suboptimal policy quickly, however, because of the system uncertainty, such as the noise from the sensors~\cite{r20}, the final learning result may not be as good as the one using sparse reward~\cite{r26}. In order to combine the advantages of both sparse and dense reward methods, in this study we propose a {Dense2Sparse} reward shaping approach. It uses a dense reward to guide the agent to learn a relative optimal policy, and then switches to the sparse reward, which utilizes the noiseless experience to further optimize the DRL algorithm.\par

In order to evaluate the performance of the {Dense2Sparse} technique, we set up a typical robot manipulation environment using a $7$ degree of freedom (DOF) robot and a monocular camera for sensing the environment. Normally, a fixed monocular camera is not able to form a stereo vision which is commonly used in the manipulation task to obtain the target's position/orientation feedback. With the help of deep learning-based image processing techniques~\cite{r27,r28,r29,r30}, DRL often takes a representation learning via transforming the images into a meaningful form with physical meaning, such as the position and distance~\cite{r1}, and this process is called state representation. In this work, we use a fixed monocular camera to extract the location of a target block by a ResNet34 network~\cite{r28}, as the state representation model to estimate state and reward signals. The experimental results show a superior performance of the usage of {Dense2Sparse}, compared with the ones using the standalone dense and sparse reward, respectively. Interestingly, as we increased the system perturbation level by shifting a certain angle of the calibrated camera (which increased the uncertainty of the feedback system), the {Dense2Sparse} approach was able to maintain almost the same performance of convergence speed and episode total rewards, which illustrates the {Dense2Sparse} is not sensitive to the system uncertainty. Therefore, the {Dense2Sparse} is able to balance the effectiveness and efficiency of DRL, especially for the robot manipulation tasks with system uncertainty. \par
The contributions of this study can be summarized as follows: (1) the proposed {Dense2Sparse} approach is able to balance the efficiency and effectiveness of DRL compared with regular dense or sparse reward shaping method. (2) The proposed {Dense2Sparse} approach is able to deal with system uncertainty, to make the DRL policy less sensitive to the noise and that helps the extension of the DRL application into more scenes.

\section{Related Work}

\subsection{Reward shaping}
The reward shaping plays an important role in the DRL as the interaction with the environment to guarantee successful operations. The efficiency and effectiveness are two major challenges of the reward shaping, which also hinder the further development of applying DRL to the robot practice directly.  \par
Most of the time, the dense reward is designed manually in various DRL settings~\cite{r5,r31}. In general, this reward calculation needs physical quantities of the object features such as position or distance information. To obtain those physical quantities, multiple sensors must be utilized which may introduce extra noise and increase the system complexity~\cite{r32,r33}. Other than dense reward, sparse reward is an easier and more straightforward way to provide feedback to the DRL, which has also been widely applied in many robotic studies. For instance, in~\cite{r6}, a binary reward which is obtained by testing whether the electronics are functional or not, is used to conduct the industrial insertion asks. In~\cite{r2}, a sparse reward is designed for judging whether the target block is lifted through the visual method, which is determined by subtracting the image before and after the task. In~\cite{r25}, the agent can get a reward only when all blocks are at their goal positions in the robotic block stacking task. The results of those studies have shown that DRL can obtain relatively good performance even with a sparse reward.\par
In general, it needs a certain expert experience to design a dense reward functional which leads to good policy performance but still be suboptimal~\cite{r13,r33,r34}. Compared to dense reward, a number of tasks are natural to specify with a sparse reward, and the policy can perform well if some tricks are introduced to overcome the initial exploration problem~\cite{r2,r6,r25}. \par

\subsection{State representation and noise filtering}
Although the dense reward has an obvious advantage of faster convergence speed, in most cases the precise physical state is not accessible, which is also common in the field of robotic manipulation. State representation learning is an effective way to estimate the physical state under this situation~\cite{r8,r9,r35}. In~\cite{r36}, the state representation model contains $8$ convolutional layers and an LSTM module, which can map the sequences of the past $4$ images and joint angles to motor velocities, gripper actions, 3D cube position and 3D gripper position. In~\cite{r30}, a 3D coordinate predictor based on VGG16~\cite{r29} is trained with the amount of domain randomization, which can keep high accuracy when the environment alters from simulation to the real scene. The state representation adopted in this paper is based on a ResNet34~\cite{r28} network structure, and to keep certain robust performance, we also introduce domain randomization tricks as in~\cite{r30}. \par

Since the state representation model has some certain estimation error~\cite{r30}, the dense reward calculated according to the estimated state also contains the perturbation, which may decrease the policy's performance. A direct way to tackle the estimation error is adopting the Kalman filter~\cite{rKalman}. However, before using Kalman filter to correct the estimation error, we should know the system dynamics (e.g. robot-environment interaction model) which is inaccessible in the model-free DRL. In~\cite{r11}, the confusion matrix analysis is proposed as the unbiased reward estimator to tackle the noisy reward problem, and this strategy leads to a good improvement of learning effectiveness. However, since this method can only deal with limited discrete situations, it is difficult to extend this method to large continuous state-space which is common in the robot study.\par

It has been found that for DRL policy, the sparse reward guided policy may outperform the one with the dense reward~\cite{r8,r26}. This is because that the sparse reward has a straight judgement about the completion of the task or not, which is hardly affected by the human factor and environment noise~\cite{r26}. Considering the fast convergence property of dense reward and the robustness of sparse reward, in this study, a novel reward shaping method named {Dense2Sparse} is developed. In this method, firstly, a dense reward which may include noise is used to learn a suboptimal strategy quickly, and then switches to the sparse reward to continue the policy learning, which guarantees a fast and robust learning performance.\par

%BACKGROUND
\section{Problem Statement and Method Overview}
In this study, we propose a novel reward shape in reinforcement learning to balance the effectiveness and efficiency of the learning for robot manipulation. Specifically, the manipulation is conducted by the following two stages. At the first stage, a ResNet34 network is used as our state representation model to estimate the location of the target object from the single view images captured by a camera. This stage is convenient for the reward shaping and the hardware setup (all we need is only one monocular camera) especially for the robot manipulation, but is inevitable to introduce estimation error. In the second stage, the estimated target location is used by a DRL algorithm based on the {Dense2Sparse} technique. Specifically, the DRL runs with noisy dense reward shape for a predefined number of episodes, and then switches to the sparse reward setting for the rest of the training process.\par
In general, the manipulation tasks can be modeled as a finite-horizon, discounted Markov Decision Process (MDP). To describe the MDP, let $S$ be the low-dimension state space which is given by the state representation model; $A$ be the action space which is construed by the joint speed command and grasping command; $\gamma$ and $T$ be the discount factor and horizon, respectively, which are set to $0.9$ and $200$ in this study. Then the reward function can be represented as the mapping $S \times A\to R$, and the state transition dynamics can be denoted as $S \times A\to S$. Finally, the policy $\pi$ can be then defined as a deterministic mapping $S \to A$. The goal of formulating the MDP is to maximize the total expected reward $G(\cdot)$ (expressed as (1)) by optimizing the policy $\pi$.

\begin{equation}
G(\pi)=E_\pi \biggl[\sum_{t=0}^T \gamma\cdot r(s_t,a_t) \biggl]
\end{equation} \par
As we adopt the model-free DRL to solve the MDP problem above, it can avoid accessing the transition dynamics. The detail training process of state representation is described in section IV, and the policy training is detailed in section V.

%Training of Representation Model
\section{Representation Model}
It is a challenge for utilizing the raw visual information to conduct the reinforcement learning for robotic manipulations directly~\cite{r8}. Instead, it is more feasible to use the low-dimension state representations as the feedback data. Normally, the state representation that contains clear physical meaning is always welcomed as it can serve both as the state and the reward shape, and thus reduce the requirements of physical sensors. Specifically, we employ a pre-trained deep learning network, ResNet34, as our state representation model to process the raw images obtained by a fixed monocular camera. The loss function of the ResNet34 network is given as follows.

\begin{equation}
\begin{aligned}
L_{\rm ResNet}=&-\sum_{m=1}^{3}(d_m - \widetilde d_m)^2
\end{aligned}
\end{equation}

where $d_m$ (m =1, 2, 3) presents the actual $x$, $y$ and $z$ value in world coordinate and $\widetilde{d_m}$ denotes the predicted $x$, $y$, $z$ value output by the ResNet34 model. \par
It has been shown that combining the random data and policy related data in the data set could lead to better policy learning performance~\cite{r8}. In this study, we collected data by running the robot arm whose actions were comprised of $40\%$ random actions and $60\%$ policy related actions. To overcome the ``sim-to-real" problem (the policy works well in one simulation environment but worse in another scene), we adopted the simple domain randomization method~\cite{r30}, in which we randomly changed the position of the camera in a small range. We totally ran $1600$ episodes with $200$ steps per episode to collect the data, and finally selected $80k$ images which is one quarter of the total data.
The evaluation metric of this representation model is the mean Euclidean distance between the ground truth and the predicted position of the target block. As shown in Fig.~1, the final error of the state representation model is about $1.4$ cm after $500$ epochs training.

\begin{figure}[h]
\centering
\includegraphics[width=0.45\textwidth]{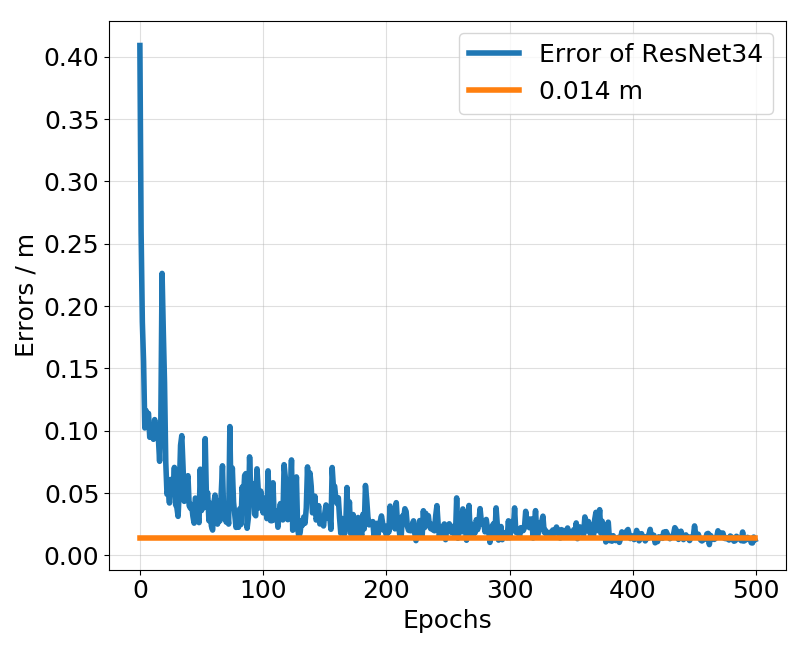}
\caption{The error graph of representation models during training.}
\label{f1}
\end{figure}

\section{Policy Learning and Experiment Setting}
\subsection{Policy learning}
We evaluate the performance of the {Dense2Sparse} strategy on two common robot manipulation tasks, reaching and lifting. To save the computing resources, both of the tasks utilize the same state representation model described in section IV. As a demonstration, we use the TD3~\cite{r16} as our reinforcement learning algorithm. The schematic diagram of the policy learning process is shown in Fig.~2. In the first stage, the state obtained from the representation model and the robot proprioception (including the angle and angular velocity of the joints as well as the opening degree of the gripper) are used to design a dense reward. The dense reward indicates that in each action step, the robot can get a reward which is calculated by the reward function where the physical quantities are obtained from the state representation model. After a certain number of episodes, the dense reward is switched to the sparse reward to continue the policy learning (the second stage). In the second stage, the robot can only get a reward when the task (reaching task) or sub-task (lifting task) is completed.

\begin{figure}[h]
\hspace*{-0mm}
\includegraphics[width=0.48\textwidth]{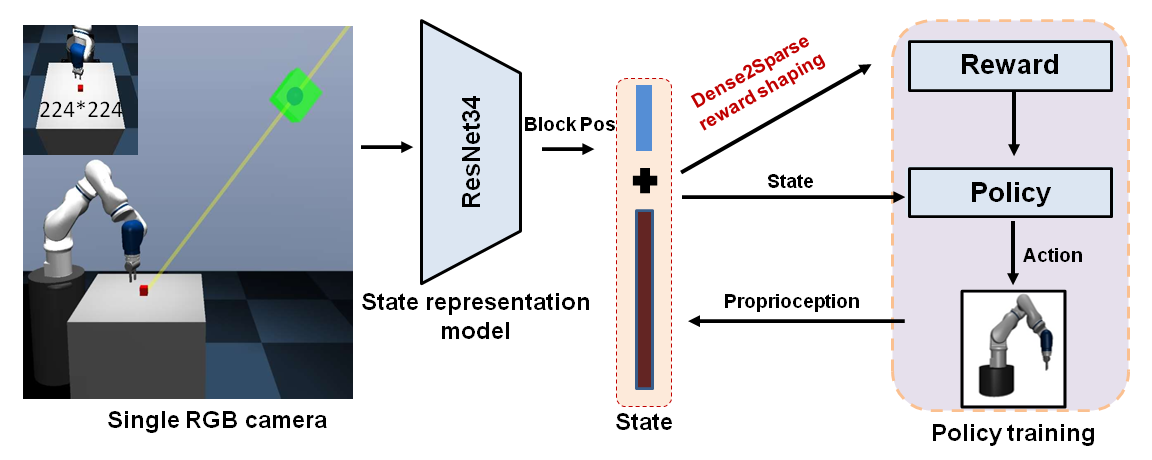}
\caption{Schematic diagram of the testing platform}
\label{Fig_ExpSetup_1}
\end{figure}

\subsection{Task setup}
Both of the reaching task and lifting task are conducted in the Robosuite~\cite{r1}, which is a simulation environment based on MUJOCO physical engine~\cite{r37}. The robot we used in this study is a 7-DOF robot (SCR5, Siasun. Co, Shenyang, China) with a two-finger gripper. For the reaching task, the gripper is kept closed, and the task is completed only when the end of the gripper touches the target block ($5\times5\times5$ cm$^3$) or within $3$ cm of the center of the block. For the lifting task, the robot has to grasp the block at first, and then lift it $4$ cm away from the desktop to complete the task.

\subsection{Reward function}
There are two types of reward functions in both the reaching task and the lifting task, sparse reward and dense reward. For the sparse reward, the robot gets a binary reward, which is $1$ if the task is completed and $0$ for others. The detail reward setting is formulated as follows. Where $d$ is the distance between the center of the gripper and the center of the block; $R_{reach-dense}$ and $R_{reach-sparse}$ represent the dense reward and the sparse reward we adopt in the reaching task; and $R_{lift-dense}$ and $R_{lift-sparse}$ represent the dense reward and sparse reward in the lifting task, respectively.

%%%%%%%%%%%%%%%%%%%%%%%%%%%%%%%%%%%%%%%%%%
\[ R_{reach-dense} =
\begin{cases}
1-tanh(10\cdot d) &  \text{if}\:d\:\:>\:0.03\:\text{m}\\ \tag{3}
1 & \text{if}\:d\:\:\le \:0.03\:\text{m}
\end{cases} \]

%%%%%%%%%%%%%%%%%%%%%%%%%%%%%%%%%%%
\[ R_{reach-sparse} =
\begin{cases}
0 &  \text{if}\:not\:touch\\ \tag{4}
1 & \text{if}\:touch
\end{cases} \]
%%%%%%%%%%%%%%%%%%%%%%%%%%%%%%%%%%%%%

\[ R_{lift-dense} =
\begin{cases}
1-tanh(10 \cdot d) &  \text{if}\ in\ reaching\ stage \\ \tag{5}
1 & \text{if}\  block \ is\ grasped \\
2.25 & \text{if}\  block \ is\ lifted
\end{cases} \]

%%%%%%%%%%%%%%%%%%%%%%%%%%%%%%%%%%%%%%%%%%%%%%%

\[ R_{lift-sparse} =
\begin{cases}
0 &  \text{if}\ in\ reaching\ stage \\ \tag{6}
1 & \text{if}\  block\ is\ touched \\
1.25 & \text{if}\ block\ is\ grasped \\
2.25 & \text{if}\  block\ is\ lifted
\end{cases} \]

%%%%%%%%%%%%%%%%%%%%%%%%%%%%%%%%%%%%%%%%%%%

\subsection{Evaluation metrics}
We evaluated the policy with the episode reward which is defined as the total reward that the agent gets in a single episode. To compare the policy with different reward shapes, we unified the evaluation reward with the dense reward described above. To ensure the fairness of the evaluation process, the episode reward is calculated by actual state (obtained from the simulation environment directly) rather than the predicted state through the state representation model.

\section{Experiment Results}
We tested the proposed scheme through a simple reaching task at first, followed by a further investigation via the lifting task. To make a comprehensive evaluation, we conducted a series of ablations in both tasks. The details of those ablations are shown as follows.

\noindent$\mathbf {Dense.}$ The experiments use $R_{reach-dense}$ or $R_{lift-dense}$ which is elaborated in section V, and keep it unchanged during the policy learning.\\
\noindent$\mathbf {Sparse.}$ The experiments use the $R_{reach-sparse}$ or $R_{lift-sparse}$ reward to conduct the entire reaching task or lifting task.\\
\noindent$\mathbf {Oracle.}$ Instead of using the state representation model, the actual state (obtained directly from the simulation environment) is employed to calculate the reward in dense. \\
\noindent$\mathbf {Dense2Sparse.}$ The experiments use the proposed method which is elaborated in Section V as the reward.\par
We design these experiments to answer the following three questions:
\\
1)	Whether the policy with the {Dense2Sparse} reward performs better than that with a standalone dense or sparse reward if the dense reward is not accurate?
\\
2)	How much performance loss if we use the {Dense2Sparse} reward compared to directly utilizing the oracle reward when there are uncertainties in state and reward?
\\
3)	How robust is our method when there is state shifting during the policy training process?

\subsection{Reaching task}
For the case with ideal camera alignment setting (as shown in Fig.~3(a)), the training curve of the TD3 agent is shown in Fig.~4(a). Every $5$ episodes have been evaluated and recorded the mean episode reward after $20$ training episodes. Each curve in Fig.~4(a) includes three replications based on $3$ random seeds. To make the results more intuitive, we also plot histograms of the success rate which is shown in Fig.~4(b).

\begin{figure}[h]
\hspace*{-1mm}
\includegraphics[width=0.48\textwidth]{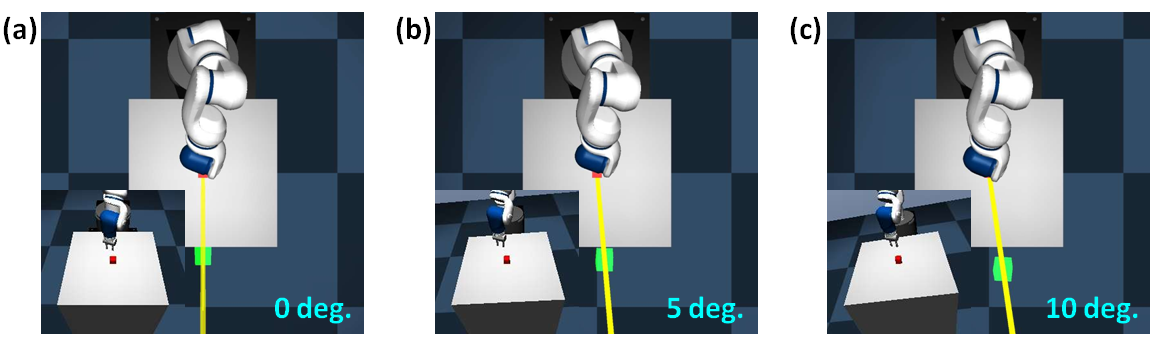}
\caption{Comparative tests with different camera setting, (a) scenario for ideal camera alignment setting, (b) scenario for $5^\circ$ camera alignment error, (c) scenario for $10^\circ$ camera alignment error.}
\label{Fig_ExpSetup_2}
\end{figure}

\begin{figure}[!htbp]
\hspace*{-1mm}
\includegraphics[width=0.49\textwidth]{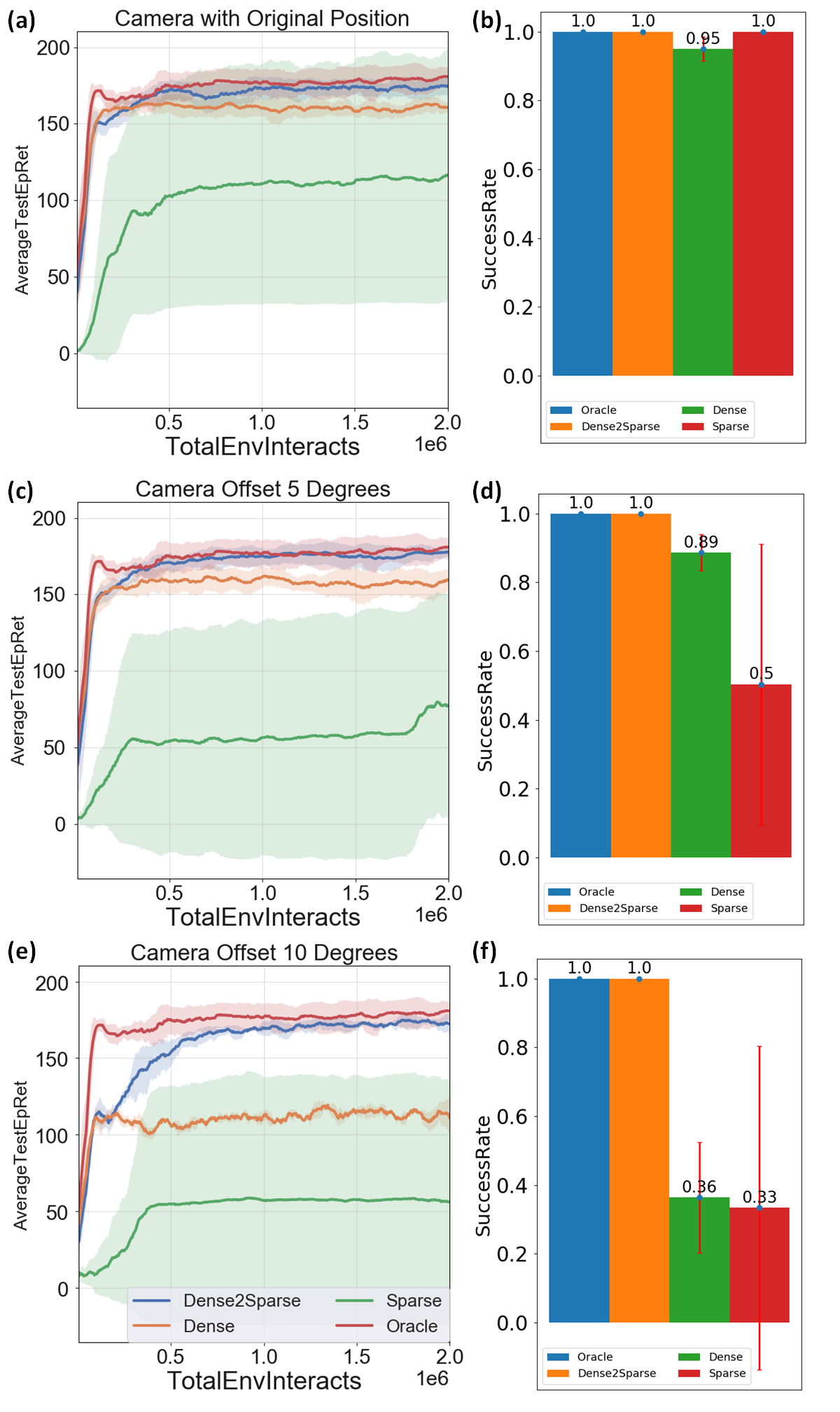}
\caption{The evaluation results in the reaching task. The solid line and transparent belt in (a), (c) and (e) represent the mean and standard deviation of 3 random seeds for three camera settings which represent no camera shifting, with $5^\circ$ camera shifting, with $10^\circ$ camera shifting, respectively. The histograms in (b), (d) and (f) present corresponding mean and standard deviation of the success rate that policy can reach after the training process is completed in (a), (c) and (e). For each evaluation, we test 1000 episodes with different target locations and robot initial positions.}
\label{Fig_BlockReaching}
\end{figure}

From Fig.~4(a) it can be seen that the policy with a standalone sparse reward has a much slower convergent speed compared to the policy with the {Dense2Sparse} reward or the standalone dense reward. The final episode reward of the policy with the standalone sparse reward can reach as high as $120$, which is much lower than other ablations. This is reasonable as the DRL policy is updated according to reward. When the reward is sparse, the policy cannot get timely reward feedback which could lead to a slower convergent speed and poor episode reward. Policy with standalone dense reward shows a similar convergence compared to the policy with {Dense2Sparse}, but its final episode reward is about $10$ less than that of the {Dense2Sparse}. An intuitive explanation is that dense reward can guide the policy to converge quickly, although error exists in the reward. However, the accumulated error of each step reduces the upper limit of the policy with a standalone dense reward. In addition, experimental result also indicates that the Oracle has a slightly higher final episode reward than the policy with {Dense2Sparse} reward. This is normal because the Oracle one has no noise at all, which guarantees the best performance of the policy. However, in practice, it is always difficult to obtain the accurate state or reward during an operation process, and in such a condition, our {Dense2Sparse} technique is likely to be the one capable of getting close to the best performance of a policy. The result in Fig.~4(b) shows that though there is much difference in the performance of episode reward among those ablations, the difference in corresponding mean success rate is not so big. The mean success rate of policy with standalone sparse reward reaches $95\%$, which is only $5\%$ lower than other ablations which can reach $100\%$. This could be due to that the reaching task is relatively simple. \par

One of the advantages of the {Dense2Sparse} strategy is to make the policy less sensitive to the noise. In order to test its performance on the noisy environment, we moved the camera with $5^\circ$ (Fig.~3(b)) and $10^\circ$ (Fig.~3(c)) shifting away from original position respectively, and other setting was kept consistent with that in the reaching task above. In such a condition, the output of the pre-trained state representation model became much more inaccurate which increased the system uncertainty. Fig.~4(c) and Fig.~4(e) show that the final episode reward of policy with the standalone sparse reward can only reach about $80$ (camera with $5^\circ$ shifting) and $55$ (camera with $10^\circ$ shifting) which is much lower than the original scene (about 120). The mean success rate also reduces to $50\%$ and $33\%$ and with a large standard deviation. It shows that with the state error going up, it is more and more difficult for the policy with sparse reward to keep the performance. The large standard deviation means that the policy we get after the training process is quite unstable. An intuitive explanation for this phenomenon is that except for the specific sparse reward, the only information that is utilized to update the policy is the state which is predicated by the state representation model. When the error of the state representation model increases, it becomes difficult to keep the policy updates following the direction of the original policy with less state error, which thus leads to worse policy performance and policy stability (high standard deviation). \par

We can see from Fig.~4 that there is also an episode reward reduction and standard deviation increasing for the policy with standalone dense reward. However, compared to the policy with standalone sparse reward, it still has a large advantage (episode reward is about $75$ and $50$ higher than the policy with standalone sparse reward when there is $5^\circ$ and $10^\circ$ degrees camera shifting). A reasonable explanation is that the dense reward can decrease the influence that caused in state error but with a limit. \par

We can also know from Fig.~4 that as for the policy with the {Dense2Sparse}, both the episode reward (175) and success rate ($100\%$) keep nearly unchanged when the noise is increasing. For the policy with {Dense2Sparse} reward, the agent only gets an inaccurate biased reward during the learning process used dense reward, but obtains an accurate reward when the task is completed after switching to the sparse reward. That is the reason why {Dense2Sparse} reward shape is naturally less sensitive to the noise. However, for the policy with standalone dense reward, the agent gets biased error every step throughout the training process without any correction, and thus its learning is not as good as the one using the {Dense2Sparse} reward shape. \par
To summarize, the testing results show that the Dense2Sparse reward shaping method has a higher convergence speed compared to the sparse reward method, which illustrates the efficiency of our method. In addition, our method is able to deal with the uncertainty of the environment, in other words, this method has a good tolerance on the noise from the observer which increases the robustness of the entire system.

\subsection{Lifting task}
In order to test if the proposed {Dense2Sparse} reward shape also works for more complicated task, the same ablative experiments were conducted in the lifting task which contains three stages (reaching, grasping and lifting), and thus it is more complicated than the previous reaching task. The experimental result of policy learning curves and the histogram of the evaluation result are shown in Fig.~5(a) and (b), respectively.\par

%\begin{figure}[htbp]
%\centering
%
%\subfigure[Training curve of the lifting task]{
%\begin{minipage}[t]{1\linewidth}
%\centering
%\includegraphics[width=2.5in]{lift1.png}
%%\caption{}
%\end{minipage}%
%}%
%
%\subfigure[Performance histogram of the lifting task]{
%\begin{minipage}[t]{1\linewidth}
%\centering
%\includegraphics[width=2.5in]{lift2.png}
%%\caption{Performance histogram of the lifting task}
%\end{minipage}%
%}%
%
%
%\centering
%\caption{evaluation results of the lifting task}
%\end{figure}

\begin{figure}[!htbp]
\hspace*{-1mm}
\includegraphics[width=0.48\textwidth]{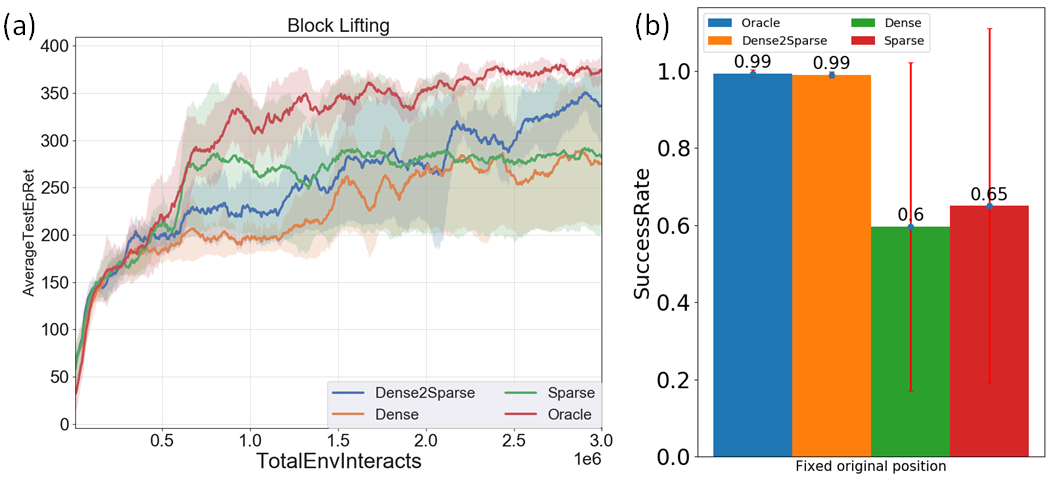}
\caption{Evaluation results of the lifting task, (a) training curve of the lifting task, (b) performance histogram of the lifting task.}
\label{Fig_BlockLifting}
\end{figure}

From Fig.~5(a), it shows that for the policy with standalone sparse reward and policy with standalone dense reward, the episode reward can reach about $260$ but do not increase anymore. For the policy with the sparse reward, it may be caused by inadequate reward feedback, and for the dense reward, it may be caused by the reward error accumulation which has been detailed in the reaching task. However, for the policy with the {Dense2Sparse} reward, the final episode reward can reach about $350$ which nearly equals to the Oracle policy. Fig.~5(b) shows that for the policy with standalone sparse reward and with standalone dense reward, the mean success rate is $60\%$ and $65\%$, respectively, which is much lower than the policy with {Dense2Sparse} reward and Oracle. Meanwhile, both the standard deviations of the policy with standalone sparse reward and with standalone dense reward are quite large, indicating that the policies are not stable which may be caused by the increasing of task complexity. It can also be seen that the mean success rate of the policy with {Dense2Sparse} can reach as high as $99\%$ and the standard deviation is nearly zero which is the same as the Oracle, which further illustrates that our method can achieve as stable performance as the Oracle. In summary, the proposed {Dense2Sparse} reward shape can tackle the state and reward uncertainty in the complex task with comparable performance as the policy using the accurate state and reward information. This illustrates the {Dense2Sparse} reward shaping method has a great potential to solve the learning effectiveness and efficiency problem in the DRL and robust to the system uncertainty in practice.

\section{Discussion and Conclusion}
In this study, we proposed a simple but effective reward shaping strategy, {Dense2Sparse} to balance the efficiency and effectiveness of the DRL for system with uncertainty. It should be noted that our {Dense2Sparse} strategy is based on the assumption that there exists system uncertainty and a clear task completion judgement (for sparse reward) is available. We believe in real practice, most of the robot manipulation scenarios belong to this category. Under such an assumption, our method is ready to combine with arbitrary off-policy DRL algorithm and shaping reward from state representation learning.\par
To verify the proposed method, we developed a stable state representation model to estimate the position of the target object, which is then used to provide a dense reward for the agent. It can guide the agent to achieve suboptimal performance, followed by switching to sparse rewards to rectify the agent behavior to get closer to optimal performance. We verified this {Dense2Sparse} approach by a series of robotic reaching and lifting tasks with system uncertainty such as camera alignment offset. The testing results show that, besides the fast convergence speed, our method also can rescue the agent from misleading rewards even at a relative high noise level. Future work will focus on combining the {Dense2Sparse} strategy with the ``sim-to-real" approach, to make the DRL more robust to the environment changes.\par
It also should be noted that we do not verify our method in the real scene as the technology of the ``sim-to-real" is quite mature such as domain randomization~\cite{r30,r38,r39}, dynamic randomization~\cite{r40,r41}, domain adaption~\cite{r42, r43} and so on, and all of above studies have already proven that if a DRL policy works in the physical engine-based simulation environment, it should work for the real practice as long as ``sim-to-real" technique has been applied.

\bibliography{IEEEabrv,reference}

\end{document}